%% file: paper.tex
\documentclass[runningheads]{llncs}
\usepackage[T1]{fontenc}
\usepackage{graphicx}

\usepackage[normalem]{ulem}
\usepackage{ulem}
\usepackage{amsmath}
\usepackage{makecell}
    
\usepackage{array} 

\usepackage{tabularx} 

\usepackage{longtable}

\usepackage{multirow}
\usepackage{booktabs}
\usepackage{hyperref}
\begin{document}
\title{From Legal Text to AI-specific Risk Sources: A Systematic Analysis of the EU AI Act's High-Risk Requirements}
\titlerunning{From Legal Text to AI-specific Risk Source}

\author{
Ronald Schnitzer\inst{1}\inst{2} \and 
Mike Auer \inst{2} \and
Rumpa Choudhury \inst{2} \and
Andreas Hapfelmeier \inst{2} \and
Maximilian Hoeving \inst{2}\and 
Isabelle Painter \inst{2} \and
Josiane Xavier Parreira \inst{2} \and
Sonja Zillner\inst{1}\inst{2}}
\institute{Technical University of Munich, School of Computation, Information and Technology, Munich, Germany \\
\email{firstname.lastname@tum.de}
\and
Siemens AG, Munich/Vienna, Germany/Austria \\
\email{firstname.lastname@siemens.com}
 }

\authorrunning{R. Schnitzer et al.}

\maketitle              
\begin{abstract}

The EU AI Act introduces mandatory requirements for high-risk AI systems with the explicit goal of ensuring the development and operation of trustworthy AI.
At the same time, AI risk management practices rely on structured risk taxonomies to systematically identify and treat AI-specific risk sources.
As both the AI Act and established risk taxonomies aim to address AI-induced risks, a natural question is whether they align in the risk sources they cover.
However, no clear mapping exists between the risks implicitly addressed by the Act's high-risk requirements and established taxonomies, leaving practitioners without a structured basis for aligning regulatory obligations with AI risk management practice.
This paper presents a systematic classification of the requirements extracted from the EU AI Act Section 2 ("Requirements for high-risk AI systems"), revealing that only a minority directly address AI-specific risk sources, while the majority impose organizational process and documentation obligations. From the AI risk-related requirements, a consolidated list of distinct AI-specific risk sources is derived. The resulting EU AI Act Risk Source List takes an important step towards bridging the gap between legal obligation and AI risk management practice, providing a structured reference for explicit comparison between existing AI risk taxonomies and the risk sources implicitly addressed by the EU AI Act.
\textbf{Important Note: This is the authors’ preprint. The paper was presented at the 4th International Conference on Frontiers of Artificial Intelligence, Ethics, and Multidisciplinary Applications. A link to the conference’s official proceedings will be provided upon publication.
}
\end{abstract}
\section{Introduction}
\label{sec: Introduction}

The EU AI Act (AIA) represents one of the world's first large-scale regulatory frameworks for Artificial Intelligence (AI) systems \cite{AI_Act}. 
Specifically, the AIA defines a four-level risk classification scheme (e.g., unacceptable, high, limited, minimal risk) to align regulatory obligations according to the risk level of AI systems. While minimal-risk and limited-risk AI systems face only a small set of legal obligations, and AI systems classified as unacceptable risk are strictly banned within the EU jurisdiction, high-risk AI systems are subject to a comprehensive set of regulatory obligations defined in Articles 8--15 of the Regulation, hereafter referred to as High-Risk Requirements (HRRs).
As the Regulation notes in Recital 64, the purpose of the HRRs is "to mitigate the risks from high-risk AI systems placed on the market or put into service and to ensure a high level of trustworthiness" \cite{AI_Act}.

Although the Regulation entered into force in August 2024, the high-risk requirements apply only after transition periods that differ by category: from 2 August 2026 for stand-alone high-risk systems listed in Annex III, and from 2 August 2027 for high-risk AI embedded in products covered by Annex I \cite{AIActTimeline}. While the HRRs are specified in the Regulation itself, the standardization organization CEN-CENELEC has been mandated to develop harmonized standards as a primary compliance pathway \cite{AIActTimeline}. At the time of writing, however, these remain a work in progress with a risk of delay, leaving organizations in a state of uncertainty \cite{cencenelec2025ai}.

The AIA's intention with defining the HRRs aligns with engineering approaches to AI risk management, which is the systematic identification, assessment and mitigation of root causes of AI-specific risks \cite{ISO_5469,NIST,schnitzer2023ai}. Previous work has been focusing on the development of AI risk taxonomies, such as the MIT AI risk repository \cite{MIT_Risk_Repository}, the AI risk atlas \cite{bagehorn2025ai} or ISO 5469 \cite{ISO_5469}, among others \cite{schnitzer2023ai,steimers2022sources,weidinger2021ethical,willers2020safety,zendel2015cv}. These taxonomies vary in their scope (e.g., GenAI, computer vision, etc.) and focus (e.g., ethics-risks, safety risks) and granularity (e.g., high-level guidance vs. detailed enumerations of failure modes).
AI risk taxonomies provide the foundation for systematic risk identification and have found adoption in industrial AI risk management and governance practices \cite{zillner2025responsible}.

While the HRRs are intended to mitigate risks posed by high-risk AI systems \cite{AI_Act}, as a legal instrument, it does not explicitly enumerate AI-specific risk sources. Instead, while some HRRs are directly driven by AI-specific risks, others cover obligations not directly linked to a risk source, such as requirements to produce documentation or establish governance processes.
For instance, Article 13.3(a) merely requires that the instructions for use contain "the identity and contact details of the provider" \cite{AI_Act}, which carries no relation to an AI-specific risk.
Because the HRRs are formulated as legal requirements rather than risk taxonomies, the AI risks that motivate them are only implicitly embedded and not explicitly stated by the Regulation itself.
To the best of our knowledge, no systematic extraction of these implicit risk sources from the AIA has been conducted to date, leaving practitioners without a structured basis for assessing whether their risk management practices cover what the AIA implicitly demands. This gap directly motivates the research questions of this study:

\begin{enumerate}
    \item What proportion of the EU AI Act's high-risk requirements directly relate to AI-specific risk sources?
    \item What are the AI-specific risk sources that are implicitly addressed in the high-risk requirements of the EU AI Act?
\end{enumerate}

To answer these research questions, we apply a requirements engineering approach grounded in ISO/IEC/IEEE 29148:2018, combined with systematic content analysis, to extract and classify the HRRs of the AIA.
The contributions of this paper are threefold:
\begin{enumerate}
    \item a systematic classification of the extracted HRRs into AI risk-related requirements (AIR), process compliance requirements (PCR), and documentation compliance requirements (DCR), providing a structured analytical framework for assessing the regulatory approach of the AIA;
    \item a derived EU AI Act Risk Source List that makes the AI-specific risk sources implicitly addressed by the high-risk requirements of the AIA (Articles 9--15) explicit, providing a structured reference for AI risk identification and assessment as well as the reference point that a coverage comparison against established AI risk taxonomies presupposes; and
    \item a transferable methodology for the systematic extraction and classification of regulatory obligations, grounded in ISO/IEC/IEEE 29148:2018 and applicable to other regulatory instruments beyond the AIA.

\end{enumerate}

The remainder of the paper is structured as follows. Section~\ref{sec:rw} provides background on the AIA and AI risk management practices relevant to this study. Section~\ref{sec:met} outlines the research approach and the employed methodology. Section~\ref{sec:results} presents the study's results. Section~\ref{sec:disc} discusses the findings, reflects on the study's implications and limitations, and Section~\ref{sec:conc} concludes the paper with directions for future work.

\section{Related Work}
\label{sec:rw}
This study investigates the relationship between the AIA HRRs and AI-specific risk sources. We organize related work into three clusters: (1) legal analyses of the AIA and its HRRs; (2) taxonomies and repositories of AI risks; and (3) structured knowledge representations of HRRs and related concepts.

Legal analyses of the AIA are essential because they delineate the scope and intent of HRRs, clarifying what must be satisfied and interpreted before any systematic classification is possible. Almada \cite{almada2025eu} offers a detailed legal examination of how product safety and fundamental rights protections are jointly pursued in the AIA, explicating the dual objectives that underpin the high-risk regime and the interpretive tensions this duality can create. Complementing this, Golpayegani et al. \cite{golpayegani2023high} investigate the conditions under which AI systems are classified as high-risk and seek to formalize the resulting concepts to support AI risk management. Together, these works illuminate the legal and conceptual foundations of high-risk categorization and the obligations attached to HRRs, providing the regulatory backdrop against which our classification operates.

Systematic taxonomies and repositories of AI risks are related to this work because they enumerate the AI-specific hazards and sources of risk that HRRs are intended to mitigate. The MIT Risk Repository \cite{MIT_Risk_Repository} synthesizes a meta-review and taxonomy of risks arising from AI, while Bagehorn et al. \cite{bagehorn2025ai} provide a taxonomy and associated tooling for navigating AI risks. Complementary perspectives include safety-engineering concerns specific to AI systems \cite{willers2020safety}, ethical risk categories for large language models \cite{weidinger2021ethical}, enumerations of risk sources in AI \cite{steimers2022sources}, and concrete lists of AI-specific hazards \cite{schnitzer2023ai}. Collectively, these resources offer comprehensive inventories and organizing principles for AI risks across technical, ethical, and application domains. They provide candidate risk sources and conceptual lenses that can inform a structured linkage between identified risks and the HRRs formulated in the AIA.

Machine-readable representations of HRRs and related risk concepts are increasingly important for analysis, traceability, and the development of compliance-support tools. The TAIR knowledge graph \cite{hernandez2025open} models HRRs and related concepts but relies on an earlier, now outdated version of the AIA. AIRO \cite{golpayegani2022airo} introduces an ontology for AI systems and risks aligned with the AIA, providing a schema without a comprehensive, systematically populated list of AI risks. VAIR \cite{golpayegani2023high} contributes a vocabulary for modeling AI risks in relation to the AIA. These efforts establish a valuable foundation for representing obligations and risks but leave open the need for a systematic, up-to-date classification that explicitly relates HRRs to AI-specific risk sources. 

In contrast to existing work, which either analyzes the AIA from a legal perspective without systematic risk source identification or develops risk taxonomies without grounding them in the AIA's specific obligations, this study bridges both through a systematic requirements engineering methodology for extracting and classifying HRRs and deriving an explicit AI Act Risk Source List as a benchmark for evaluating the coverage of existing risk taxonomies.

\section{Methodology}
\label{sec:met}

\begin{figure}
    \centering
    \includegraphics[width=1\linewidth]{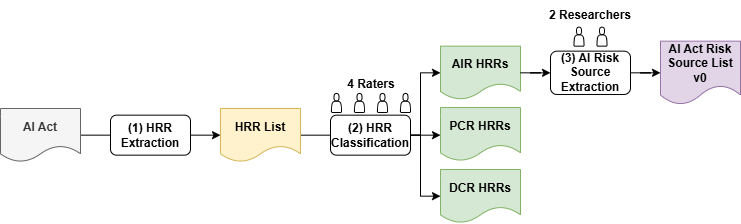}
    \caption{Three-step research process for extracting AI-specific risk sources from the EU AI Act.}
    \label{fig:methodology}
\end{figure}

The research process consists of three steps illustrated in Figure~\ref{fig:methodology} and described in the following subsections.

\subsection{HRR Extraction}

The classification of HRRs requires a list of such requirements, which the raw legal text (e.g., Section 2 of the AIA) does not provide: Articles 8--15 are formulated as continuous legal prose rather than as an enumerated set of requirements.
Extracting the HRRs from this text therefore forms the foundation of all subsequent quantitative analysis.

The extraction followed a systematic linguistic decomposition grounded in ISO/IEC/IEEE 29148:2018 \cite{ISO_29148}. According to the standard, the formulation of requirements in natural language consists of functional components, such as the \textit{Condition} under which the requirement holds, the \textit{Subject} to whom it applies, the \textit{Modal Verb} indicating the degree of obligation, the \textit{Action} to be performed, and the \textit{Constraint of Action} specifying the circumstances under which the action must be executed.
The normative statements in the AIA's text largely conform to this structure. 
However, a systematic deviation was observed: the legislative text frequently employs enumerations, specifying high-level requirements in more detailed sub-requirements without restating shared components, such as the Subject and Modal Verb for each point.
Following ISO/IEC/IEEE 29148:2018, each enumerated point was treated as inheriting shared components from its parent statement, thereby yielding distinct requirements even when those components were not repeated verbatim.

The unit of analysis was derived from the AIA's own hierarchical organization, which reflects the Regulation's logical structuring of obligations:
the Articles (8--15) are divided into numbered paragraphs (1, 2, …), which are in turn subdivided into points denoted by literals ((a), (b), …) and, in some cases, into sub-points denoted by roman numerals ((i), (ii), …).
Taking the smallest addressable structural unit, e.g., Art. 9.5(a), as one candidate requirement yields a first decomposition. 
Examining the text revealed that this purely structural criterion does not yet produce atomic requirements, for two reasons. First, a single structural unit may comprise several blocks, structurally highlighted by empty lines in the AIA, each carrying its own obligation (e.g., Art. 15.5 consists of three such blocks); 
these were split into the corresponding number of units.
Second, even individual blocks occasionally contain multiple modal verbs and thus express several distinct obligations (e.g., Art. 9.6: "High-risk AI systems shall be tested […]. Testing shall ensure that […]."). Modal verbs (shall, must, may) served as reliable delimiters for two reasons: they carry specific normative semantics in standardization contexts, distinguishing mandatory requirements from permissions \cite{iso2016howtowrite}, and each typically introduces a complete requirement structure (Subject, Action, Condition) as defined by ISO/IEC/IEEE 29148:2018.
The above-described procedure yielded 111 candidate HRRs.

While these candidates were obtained by a purely structural and lexical procedure, examining their content showed that not all of them constitute requirements in substance.
Four exclusion criteria were therefore applied sequentially.
EXC1 removes Article 8, which imposes no obligation of its own but merely stipulates that the requirements of the following articles shall be met (N=4).
EXC2 removes statements whose modal verb is may, retaining only shall and must, as the former indicates permissions rather than mandatory obligations \cite{iso2016howtowrite} (N=7). 
EXC3 removes statements whose subject is the European Commission, Notified Bodies, or parts of the AIA itself, as these concern delegated regulatory powers rather than obligations imposed on providers or deployers of AI systems (N=6).
EXC4 removes statements that merely scope the requirements following them without imposing an obligation of their own
(e.g., Art. 9.5, second sentence: "In identifying the most appropriate risk management measures, the following shall be ensured:", where the substantive requirements are given in Art. 9.5(a)--(c)) (N=7).
Their sequential application reduces the candidate set from 111 to 107, 100, 94 and finally 87 HRRs, which form the foundation for the analysis in the subsequent steps.

Since every step is anchored in the Regulation's own structure, in explicit lexical markers, or in the exclusion criteria stated above, the decomposition is rule-based and re-derivable: applying the same rules to the same text yields the same set of 87 HRRs.
\subsection{HRR Classification}
\label{subsec:HRRClassification}
To identify which HRRs are related to AI-specific risk sources, we apply a classification procedure to all HRRs extracted in the previous step.
As mentioned earlier, an initial examination of the extracted requirements reveals that not all 87 obligations relate to AI-specific risk sources. For instance, Article 9.1 “A risk management system shall be established, implemented, documented and maintained in relation to high-risk AI systems” \cite{AI_Act}, imposes an organizational governance obligation that has direct equivalents in general risk management standards such as ISO 31000 \cite{ISO_31000}, and does not address any risk source specific to AI systems.
This observation reflects a fundamental characteristic of the AIA’s regulatory architecture: like other EU product-safety-style regimes, it combines (1) obligations that target the technical and behavioral properties of systems where risk sources manifest, (2) obligations that prescribe how providers must organize and operate governance processes, and (3) obligations that require evidence sufficient to demonstrate conformity within a formal assessment regime \cite{almada2025eu,buscemi2025assessing,golpayegani2023high}.

This structure provides the theoretical basis for a three-category classification task.
Accordingly, we define the following categories: AI risk-related Requirements \textbf{(AIR)}, which directly address AI-specific risks and their sources; Process Compliance Requirements \textbf{(PCR)}, which mandate the establishment or execution of a governance or management process; and Documentation Compliance Requirements \textbf{(DCR)}, which require the production, maintenance, or disclosure of a specific artifact. Notably, the classification task is not modeled as mutually exclusive, i.e., an atomic HRR may be classified into more than one category. For instance, the HRR reflecting Article 9.1 ("A risk management system shall be established, implemented, documented, and maintained in relation to high-risk AI systems") illustrates this point well. While the \textit{requirement to establish a risk management system} points toward a classification as a PCR, the \textit{explicit mention of documentation} within the same obligation also suggests a classification as a DCR.

To operationalize this classification, a codebook was developed grounded in the classification logic described above and in Krippendorff's content analysis methodology \cite{krippendorff2018content}. 
Each category is explained in detail, including differentiation criteria and decision guidelines.

Four independent raters with backgrounds in AI engineering and AI research then applied the classification task, a profile considered appropriate given that the AIR category requires domain knowledge to identify AI-specific risk sources, but one that also has limitations, as outlined in Section~\ref{sec:limitations}.

To reflect that HRRs may be attributed to more than one category, raters were permitted to assign a secondary classification alongside their primary classification, where applicable.
To ensure the codebook was fully comprehended by all raters, an initial classification task was performed on a pilot subset of 21 requirements, three from each of the seven articles, representing approximately 24.14\% of the full dataset. 
This subset was deemed necessary given the variation in regulatory content across articles and the complexity of classifying HRRs into these categories.
The codebook was iteratively refined through successive rounds of independent coding of the pilot subset, reconciliation, and revision until Fleiss' Kappa reached or exceeded 0.61 for all three categories individually, and the average pairwise Cohen's Kappa exceeded 0.61 across all rater pairs, indicating substantial agreement as defined by Landis \& Koch \cite{landis1977measurement}. Once this threshold was achieved, raters proceeded to classify the full dataset of 87 requirements.

To obtain a final class based on the independent annotations from the four raters, a majority vote was used. However, a known limitation of a straightforward majority vote is that it does not account for raters' confidence in each classification \cite{davani2022dealing} and does not reflect the multiple-annotation setup used in this study. 
To account for this, raters were asked to provide a confidence score for each label assigned, using a three-point ordinal scale, e.g., Low, Medium, and High. Following completion of the full labeling task, the majority vote was determined by weighting each annotation by the confidence score, with higher-confidence labels carrying proportionally greater influence. 
In particular, the following calculation was applied for each requirement $HRR_i \in \{HRR_1,...,HRR_{87}\}$, rater $R_j \in \{R_1,R_2,R_3,R_4\}$, and category $c \in \{\text{AIR}, \text{PCR}, \text{DCR}\}$. 
Let $r_{c,j}(HRR_i) \in \{0,1\}$ be the binary rating for category $c$ provided by rater $R_j$ for $HRR_i$, and $w_{i,j} \in \{0.33 \text{=Low}, 0.66\text{=Medium}, 1.00\text{=High}\}$ the weight corresponding to rater $R_j$'s confidence level in rating $HRR_i$. Then for each category the category score is determined by:

\begin{equation}
    \sigma_c(HRR_i) = \sum_{k=1}^{4} w_{i,k} \cdot r_{c,k}(HRR_i)
\end{equation}

The final class is assigned by:

\begin{equation}
    \hat{y}(HRR_i) = \underset{c}{\arg\max} \ \sigma_c(HRR_i)
\label{eq:argmax}
\end{equation}

In cases where the weighted vote resulted in a tie between two categories, e.g., the argmax in Equation \ref{eq:argmax} is not uniquely determinable, the affected requirements were returned to the raters for a final consensual decision.

\subsection{AI Risk Source Extraction}
\label{sec:met_3}
To derive the AI-specific risk sources addressed by the HRRs, the four raters were prompted to provide a free-text risk source label for each requirement they annotated as AIR during the classification task described above. For example, one label provided for the HRR \textit{"Training, validation and testing data sets shall be relevant, sufficiently representative, and to the best extent possible, free of errors and complete in view of the intended purpose."} (Art. 10.3) was \textit{"Incomplete data"}. These labels served as the basis for deriving the EU AI Act Risk Source List, to which a two-researcher independent synthesis procedure was applied:

For each HRR, both researchers independently read the requirement text alongside the free-text labels provided by the four raters. Notably, the free-text labels were presented to both researchers without any information indicating which of the four raters had provided each label. Labels deemed semantically equivalent, referring to the same underlying AI risk source regardless of surface-level phrasing differences, were merged into a single representative label.
To minimize interpretative bias of both the raters who provided the initial labels and the researchers performing the synthesis task, only labels directly traceable to the HRR text were retained.
Where the requirement text implied more than one distinct risk source, multiple labels were permitted.

Following independent synthesis, both researchers compared their outputs per requirement in a structured reconciliation session. Cases of full agreement were accepted directly as final. Cases of partial or full disagreement were resolved through negotiated agreement \cite{campbell2013coding,o2020intercoder}, which is a procedure in which researchers discuss discrepant judgments until consensus is reached, with the requirement text serving as the primary arbiter throughout. The resulting EU AI Act Risk Source List, together with its mapping to the corresponding HRRs, is presented in Section~\ref{sec:results}.

\section{Results}
\label{sec:results}

The following section presents the results of the three-step research process described in Section~\ref{sec:met}. First, the classification results are reported alongside inter-rater reliability metrics and the distribution across Articles 9--15, providing the empirical basis for answering RQ1. Second, the derived EU AI Act Risk Source List is presented, providing the foundation for answering RQ2.

\subsection{HRR Classification} 

\subsubsection{Inter-rater Reliability}

Each of the four raters independently classified all 87 requirements.
Table~\ref{tab:kappa_matrix_category} reports the inter-rater reliability metrics achieved on the full dataset.

\input{Tables/IRR_1}

Overall, the raters demonstrated substantial agreement with a Fleiss' $\kappa$ of 0.69, treating the classification as a three-class nominal task. Per-category analysis revealed Fleiss' $\kappa$ values of 0.61 for AIR, 0.66 for PCR, and 0.73 for DCR, all meeting the threshold for substantial agreement (i.e., 0.61) as defined by Landis \& Koch \cite{landis1977measurement}.
DCR yielded the highest agreement, consistent with its relatively unambiguous documentary obligations, while AIR yielded the lowest, reflecting the higher degree of domain judgment required to identify AI-specific risk sources. Pairwise Cohen's $\kappa$ values ranged from 0.46 to 0.83 across all rater pairs and categories. Although some pairwise values fall below the threshold of 0.61, the overall Fleiss' $\kappa$ values across all categories nonetheless meet the threshold for substantial agreement.
It should be emphasized, however, that agreement is weakest for AIR, the category on which both research questions rest, and that agreement among raters of a shared professional profile is a measure of consistency rather than of interpretive validity.
These aspects are discussed in Section~\ref{sec:limitations}.

\subsubsection{Distribution of Categories}

The classification of the 87 HRRs yielded 27 AIR requirements (31.03\%), 40 PCR requirements (45.98\%), and 20 DCR requirements (22.99\%). Table~\ref{tab:class_distribution} illustrates the distribution of categories across Articles 9--15.

\input{Tables/Article_Classification}

The results indicate a structured distribution of categories across the articles, suggesting different roles of the provisions within the regulatory framework for high-risk AI systems.

\textbf{Article 9, Risk management system} contains 18 requirements, all classified as PCR, establishing the organizational risk management framework that mirrors established standards such as ISO 31000 \cite{ISO_31000} and ISO Guide 51 \cite{ISO_Guide_51}.
While Article 9 explicitly requires providers to identify and mitigate risks to health, safety, and fundamental rights, which encompass both AI-specific and non-AI-specific risk sources, the PCR classification reflects the nature of the obligation rather than an absence of risk-relevant content.

\textbf{Article 10, Data and data governance} contains 19 requirements, of which 11 were classified as AIR. This article has the highest number of AIR requirements across all articles, consistent with its focus on data-induced risks that are typical to AI engineering, including bias in training data, data gaps, and data quality deficiencies.
The remaining requirements address data governance processes (7 PCR) and one data documentation obligation (1 DCR).

\textbf{Article 11, Technical documentation} contains 6 requirements, which are evenly classified between DCR and PCR. This article primarily concerns the evidentiary obligations of providers and does not introduce requirements specifically targeting AI-related risk sources.
Its requirements concern the production and maintenance of technical documentation artifacts, as well as the processes by which they are prepared and kept current, explaining the presence of HRRs classified as DCR and PCR in this article.

\textbf{Article 12, Record-keeping} contains 8 requirements, almost exclusively classified as PCR.
While the article establishes logging and traceability obligations that may superficially suggest a documentation character, the primary obligations concern the design and operation of logging capabilities rather than the production of specific artifacts: high-risk AI systems shall \textit{technically allow for} automatic recording, and logging capabilities shall \textit{enable} the recording of relevant events.
Following the decision procedure, the dominant obligation is to establish and maintain a technical capability, which was rated as a PCR.
It should be noted, however, that the scope of what must be logged is partly risk-dependent, i.e., Art. 12.2 requires logging capabilities to enable the identification of situations that may result in the system presenting a risk, establishing an indirect link between these process obligations and the risk management framework of Article 9. 
It is this indirect link that explains why a traceability-related risk source derived from Art. 12.2 is part of the risk source list presented in Section~\ref{sec:results}, even though no requirement of Article 12 received AIR as its primary classification.

\textbf{Article 13, Transparency and provision of information to deployers} contains 16 requirements, of which 2 were classified as AIR. The AIR requirements in this article relate to human-AI interaction risks, specifically the obligation to disclose foreseeable circumstances under which the system may pose risks, and to communicate system limitations that could contribute to misuse or over-reliance. 
The remaining requirements were classified as DCR, since their focus lies on specifying the content of the \textit{instructions for use}.

\textbf{Article 14, Human oversight} contains 12 requirements, of which 8 were classified as AIR. 
This article introduces the highest concentration of AIR requirements related to human-AI interaction, including automation bias and insufficient human intervention mechanisms.
The remaining requirements were all classified as PCR and concerned the organizational and technical measures providers must implement to enable effective human oversight.

\textbf{Article 15, Accuracy, robustness and cybersecurity} contains 8 requirements, classified across all three categories (6 AIR, 1 PCR, 1 DCR).
This article addresses the broadest range of AI-specific risk sources, including lack of robustness, data poisoning and self-reinforcing bias
through feedback loops, alongside the process and documentation obligations that support these controls.

Overall, AIR requirements are concentrated in Articles 10, 13, 14, and 15, reflecting the articles that most directly address the technical and behavioral properties of AI systems, while Articles 9, 11, and 12 focus on process and documentation compliance requirements (PCR and DCR).

\subsection{Derivation of the AI-Specific Risk Source List}
\input{Tables/AIRL}
In total, 107 risk source labels were collected from the four raters across 44 HRRs.
These include all 27 HRRs ultimately classified as AIR, as well as HRRs for which at least one rater assigned an AIR classification but whose confidence-weighted majority vote yielded a different final class. 
The latter were retained to ensure comprehensive coverage of potentially AI risk-relevant obligations. Following the synthesis procedure described in Section~\ref{sec:met_3}, this process yielded a final AI Act Risk Source List comprising 40 distinct AI-specific risk sources, presented in Table~\ref{tab:AIRL} along with their mappings to the source articles of the AIA.
An interpretation of the risk source distribution, including a thematic clustering into three categories, is provided in Section~\ref{sec:discussion_interpretation}.

\section{Discussion}
\label{sec:disc}
The following discussion interprets the classification results and the derived EU AI Act Risk Source List in light of the study's research questions, reflects on the implications for research and practice, and acknowledges the study's limitations.

\subsection{Interpretation of Classification Results and AI-Specific Risk Sources}
\label{sec:discussion_interpretation}
As reported in Section~\ref{sec:results}, nearly a third of requirements directly relate to AI-specific risk sources (AIR), while the majority constitute process and documentation obligations (PCR and DCR), thereby answering RQ1.
This finding is consistent with a broader pattern in EU product safety regulation  under the New Legislative Framework, whereby legislation establishes essential  (product-related) requirements  \cite{Machinery_Regulation}\footnote{See in particular Annex~III of \cite{Machinery_Regulation}.} as well as process-based and documentation-oriented requirements, while the technical specifications of the  constantly evolving state of the art cannot be meaningfully established in  legislative text. 
Instead, the task of developing standards to fill these abstract  legal requirements with detailed specifications is delegated to standardisation  organisations via standardisation 
requests \cite{EuropeanCommission2023AIStandardisation}.

This finding is structurally significant: Article 9 functions as a meta-requirement for the entire Chapter, mandating the establishment of a risk management system that identifies and addresses all known and foreseeable AI risks throughout the system lifecycle. The absence of any AIR requirement in Article 9 indicates that the AIA deliberately delegates the identification of specific AI risk sources to providers rather than enumerating them legislatively. The AIR requirements in the subsequent articles, therefore, represent only the minimum set of AI-specific risk sources explicitly mandated by the AIA. The full scope of risks that providers must address under Article 9 may significantly exceed this set.

The 27 AIR requirements are not evenly distributed but concentrated in specific articles, reflecting three thematic clusters of AI-specific risk that the HRRs address explicitly. The first and largest cluster concerns data quality and data management-related risk sources, concentrated in Article 10, which contains the highest number of AIR requirements across all articles (11 out of 19). This cluster includes risk sources such as bias in training data, inappropriate statistical properties in the data, and inadequate data governance practices. The prominence of this cluster is consistent with the role of data quality as one of the primary sources of AI system failures and discriminatory outcomes.

The second cluster addresses human-AI interaction risk sources, distributed across Articles 13 and 14, which together account for 10 AIR requirements.
This cluster includes automation bias, which refers to the tendency of humans to over-rely on AI-generated outputs, as well as risks arising from insufficient transparency about system capabilities and limitations, and the potential for misuse under foreseeable conditions.
The distribution of these risk sources across the transparency and human oversight articles reflects the AIA's recognition that AI-specific risks do not arise solely from technical system properties but also from the cognitive and behavioral dynamics of human-AI interaction.

The third cluster concerns AI system-specific technical risks, introduced primarily in Article 15, which contains 6 AIR requirements.
This cluster includes AI-related risk sources, such as lack of robustness, data poisoning, and self-reinforcing bias through feedback loops.
Notably, Article 15 addresses these risk sources at a comparatively high level of abstraction, mandating appropriate levels of accuracy, robustness, and cybersecurity.
While the HRRs explicitly address some cybersecurity-related AI-specific risks, such as data poisoning or model evasion, the requirements for accuracy or robustness are not specified with the same level of detail. This indicates a legislative choice to leave the technical operationalization of these concepts to harmonized standards and implementing acts.

Taken together, the three clusters suggest that the AIA's explicit attention to AI-specific risk sources is selective rather than comprehensive, with risk sources related to model explainability, distribution shift, and emergent behavior not explicitly addressed in the AIR requirements, although they are typically covered by existing AI risk taxonomies. The EU AI Act Risk Source List, comprising 40 distinct AI-specific risk sources derived through the research process described in Section~\ref{sec:met}, represents the study's answer to RQ2.

\subsection{Implications for Research}
The methodology developed in this study, combining systematic requirements extraction, classification of regulatory obligations, and structured synthesis, provides a transferable approach for systematically analyzing AI regulation. The procedure could be applied to other regulatory instruments, such as AI regulations from other jurisdictions or the Cyber Resilience Act \cite{CRA}, enabling cross-jurisdictional comparison of regulatory risk coverage. 
The three-category classification scheme (AIR, PCR, DCR) provides a reusable analytical framework that may be applied to other technology regulations to assess their regulatory approach. When applied to other domains, the AIR category would need to be redefined to reflect the domain-specific risks of the respective regulatory context.

A formal coverage analysis against established AI risk taxonomies is deliberately left to future work rather than included here. Such a comparison presupposes an explicit, provision-traceable account of the risk sources the AIA itself mandates, which the Regulation does not provide and which this study establishes; without such a reference point, both the choice of taxonomy to compare against and the granularity at which entries are matched would remain arbitrary. The list presented here, therefore, represents a necessary first step towards that comparison rather than a substitute for it.

\subsection{Implications for Practice}
The derived list of AI-specific risk sources provides practitioners with an actionable first reference for AI risk management under the AIA. Unlike the legislative text itself, which distributes risk-relevant content across multiple articles and embeds it within process and documentation obligations, the synthesized list consolidates all explicitly mandated AI-specific risk sources into a single structured inventory.
This enables compliance teams and AI engineers to directly identify the minimum set of AI-specific risk sources to be addressed under the AIA, and to structure their risk assessment activities accordingly. It must be emphasized that the list is a minimum set and not an exhaustive one: Article 9 obliges providers to identify all known and foreseeable risks of their system, so risk sources beyond those explicitly named in the AIA will have to be addressed as well. The list is therefore a starting point for risk identification rather than a compliance checklist, and we encourage practitioners to extend it based on their specific system context, intended purpose, and deployment environment.

\subsection{Limitations}
\label{sec:limitations}
Several limitations of this study should be acknowledged. 

First, the four raters share similar professional backgrounds in AI engineering and AI research and are affiliated with the same institution.
This homogeneity plausibly explains the consistency observed, but agreement within such a group shows that the codebook can be applied reproducibly rather than that its application is correct.
The qualification bears most on AIR, for which agreement was lowest (Fleiss' $\kappa = 0.61$, pairwise minimum 0.46) and on which both research questions rest.
Future studies should examine whether the results hold across more raters, including those with legal or domain-specific expertise, and in other institutional settings.

Second, the analysis is limited in scope to Section 2 of the AIA (of which Article 8 was excluded as it imposes no obligation of its own, see Section~\ref{sec:met}), selected for their direct relevance to the technical and organizational requirements of high-risk AI systems.
Additional AI-specific risk sources may be found in other parts of the Regulation, including the prohibited practices in Article 5, the general-purpose AI model provisions, and the recitals, which do not follow the same normative requirement structure as Articles 9--15 and were therefore excluded from this analysis. Nevertheless, Articles 9--15 represent the core normative obligations for high-risk AI systems, and the derived list provides a comprehensive coverage of the risk sources directly actionable within this scope.

Third, the requirements engineering approach adopted in this study involved decomposing the legal text into atomic requirements and systematically deriving AI risk sources from these. However, given that the analyzed text constitutes a section of a regulatory document, not all HRRs are specified at the same level of detail, and an implicit hierarchy exists among them. 
This structural characteristic of the source material is reflected in the resulting list.
For instance, "Vulnerability to cybersecurity-related attacks" is relatively high-level, as it is derived from the introductory paragraph of Article 15, which establishes the general principle. More granular risk sources, such as "Model poisoning", are grounded in the concrete attack scenarios described later in the same Article. 
While this mirrors the structure of Articles 9--15, it means the derived list does not yet capture the hierarchical relationships among risk sources. Our aim was to produce a comprehensive list; future work may focus on deriving and formalizing such hierarchical structures explicitly.

Fourth, the analysis represents a static interpretation of the AIA as originally published. Upcoming official documents \footnote{This work reflects the status as of May, 2026. By the time of publication, some of these instruments may have been released or may have progressed in their development.}, such as the Digital Omnibus on AI Regulation, following the EU Commission’s proposal \cite{Omnibus}, CEN-CENELEC Harmonized Standards, or other guidance documents issued by the AI Office, may significantly refine or extend the interpretation of individual requirements, potentially adding or modifying the set of AI-specific risk sources that providers must address. 
 The list should be updated as these instruments are developed.

\section{Conclusion}
\label{sec:conc}
This paper presented a systematic classification of the HRRs of the AIA (Articles 9--15), revealing that these primarily rely on organizational governance and documentation obligations rather than direct technical controls for AI-specific risk sources. Only 31.03\% of the 87 extracted requirements directly address AI-specific risk sources, concentrated in Articles 10, 13, 14, and 15.
From these requirements, 40 distinct AI-specific risk sources were derived and consolidated into three thematic clusters: data quality and management risks, human-AI interaction risks, and AI system technical risks. 
Analogous to hazard taxonomies in established AI risk management practice, the resulting EU AI Act Risk Source List provides practitioners with a structured inventory of the risk sources Articles 9--15 address, taking a first step in bridging the gap between legal obligation and engineering practice. The list represents a minimum set of AI-specific risk sources to be considered in AI risk management, as Article 9 requires providers to identify all foreseeable AI risks beyond those explicitly named, and should be regarded as a version 0 subject to refinement as implementing acts and harmonized standards evolve.
Future work should extend the analysis to the full AIA, replicate the classification with raters from diverse professional backgrounds to validate the AIR category assignments, apply the methodology to other regulatory instruments, and conduct a formal coverage analysis against existing AI risk taxonomies, including the "AI Hazard List" \cite{schnitzer2023ai}, the MIT Risk Repository \cite{MIT_Risk_Repository}, and the AI Risk Atlas \cite{bagehorn2025ai}. Alignment with existing ontologies such as TAIR \cite{hernandez2025open} and AIRO 
\cite{golpayegani2022airo} should also be investigated.

\subsection*{Acknowledgements}
This project is supported by funds from the European Union’s ’DNS of FutureProof Mobility. Digital – Sustainable – System-Compatible’ program of the Federal Ministry for Economic Affairs and Energy (BMWE).

\bibliographystyle{splncs04}
\bibliography{bib}

\end{document}

%% file: Tables/IRR_1.tex
\begin{table}[h]
\centering
\caption{Pairwise Cohen's $\kappa$ (upper triangle) and Fleiss' $\kappa$ per category modeled as binary classification tasks (AIR,PCR,DCR), and for the overall primary category assignment modeled as a three-class nominal classification task.}
\label{tab:kappa_matrix_category}
\begin{tabular}{llcccc}
\toprule
\textbf{Category} & & \textbf{Rater 1} & \textbf{Rater 2} & \textbf{Rater 3} & \textbf{Rater 4}\\
\midrule
\multirow{4}{*}{\makecell[c]{\textbf{AIR} \\ \textbf{(Fleiss $\kappa = 0.61$)}}}
 & \textbf{Rater 1} & ---  & 0.64 & 0.70 & 0.65\\
 & \textbf{Rater 2} &      & ---  & 0.46 & 0.51\\
 & \textbf{Rater 3} &      &      & ---  & 0.77\\
 & \textbf{Rater 4} &      &      &      & --- \\
\midrule
\multirow{4}{*}{\makecell[c]{\textbf{PCR} \\ \textbf{(Fleiss $\kappa = 0.66$)}}}
 & \textbf{Rater 1} & ---  & 0.68 & 0.68 & 0.63\\
 & \textbf{Rater 2} &      & ---  & 0.59 & 0.54\\
 & \textbf{Rater 3} &      &      & ---  & 0.82\\
 & \textbf{Rater 4} &      &      &      & --- \\
\midrule
\multirow{4}{*}{\makecell[c]{\textbf{DCR} \\ \textbf{(Fleiss $\kappa = 0.73$)}}}
 & \textbf{Rater 1} & ---  & 0.71 & 0.75 & 0.78\\
 & \textbf{Rater 2} &      & ---  & 0.77 & 0.58\\
 & \textbf{Rater 3} &      &      & ---  & 0.78\\
 & \textbf{Rater 4} &      &      &      & --- \\
\midrule
\midrule
\multirow{4}{*}{\makecell[c]{\textbf{Primary Category} \\ \textbf{Assignment} \\ \textbf{(Fleiss $\kappa = 0.69$)}}}
 & \textbf{Rater 1} & ---  & 0.74 & 0.64 & 0.69\\
 & \textbf{Rater 2} &      & ---  & 0.66 & 0.61\\
 & \textbf{Rater 3} &      &      & ---  & 0.83\\
 & \textbf{Rater 4} &      &      &      & --- \\
\bottomrule
\end{tabular}
\end{table}

%% file: Tables/Article_Classification.tex
\begin{table}[htbp]
  \centering
  \renewcommand{\arraystretch}{1.20}
  \caption{The distribution of classified HRRs across the Articles 9-15 of the AIA.}
  \label{tab:class_distribution}
    \begin{tabularx}{\textwidth}{>{\raggedright\arraybackslash\hsize=2.947\hsize}X c >{\centering\arraybackslash\hsize=0.368\hsize}X >{\centering\arraybackslash\hsize=0.337\hsize}X >{\centering\arraybackslash\hsize=0.348\hsize}X}
      \toprule
      \textbf{Article} & \textbf{\# HRRs} & \textbf{AIR} & \textbf{PCR} & \textbf{DCR} \\
      \midrule
      Art. 9: Risk Management & 18 & 0 & 18 & 0 \\
      Art. 10: Data and data governance & 19 & 11 & 7 & 1 \\
      Art. 11: Technical documentation & 6 & 0 & 3 & 3 \\
      Art. 12: Record-keeping & 8 & 0 & 7 & 1 \\
      Art. 13: Transparency and provision of information to deployers & 16 & 2 & 0 & 14 \\
      Art. 14: Human oversight & 12 & 8 & 4 & 0 \\
      Art. 15: Accuracy, robustness and cybersecurity & 8 & 6 & 1 & 1 \\
      \midrule
      Sum & 87 & 27 & 40 & 20 \\
      Percentage[\%] & — & 31.03 & 45.98 & 22.99 \\
      \bottomrule
    \end{tabularx}
\end{table}

%% file: Tables/AIRL.tex
\begin{table}[htbp]
  \centering
  \caption{EU AI Act Risk Source List (v0): the minimum set of AI-specific risk sources implicitly addressed by Articles 9--15 of the EU AI Act, with mapping to source articles, distributed across three clusters (see Section \ref{sec:discussion_interpretation}).}
  \label{tab:AIRL}
    \begin{tabularx}{\textwidth}{c >{\raggedright\arraybackslash\hsize=1.434\hsize}X >{\raggedright\arraybackslash\hsize=0.6015\hsize}X | c >{\raggedright\arraybackslash\hsize=1.363\hsize}X >{\raggedright\arraybackslash\hsize=0.6015\hsize}X}
      \toprule
      \textbf{ID} & \textbf{AI Risk Source} & \textbf{Arts.} & \textbf{ID} & \textbf{AI Risk Source} & \textbf{Arts.} \\
      \midrule
      \multicolumn{3}{l|}{\textbf{Data and Data Governance}} &21 & Harmful model bias & §13.3(b)(v)\\
      \cmidrule{1-3}
      1 & Inappropriate data collection process & §10.2(b) & 22 & Inappropriate hardware for AI system operation & §13.3(e) \\
      2 & Inappropriate data source & §10.2(b) & 23 & Inappropriate human oversight set-up & §14.1 \\
      3 & Inaccurate data labels & §10.2(c) & 24 & Inappropriate human AI system interfaces & §14.1 \\
      4 & Inappropriate data preparation & §10.2, §10.3 (1), §10.3 (2), §10.4 & 25 & Inappropriate or non-proportionate monitoring options & §14.4(a) \\
      5 & Insufficiently representative data & §10.2(d) & 26 & Insufficient understanding of AI system by human oversight personnel & §14.4(a) \\
      6 & Inappropriate data quantity & §10.2(e) & 27 & Automation bias & §14.4(b) \\
      7 & Inappropriate data suitability for intended purpose & §10.2(e) & 28 & Insufficient skill/knowledge of human oversight personnel & §14.4(c) \\
      8 & Insufficient data availability & §10.2(e) & 29 & Unavailability of intervention functionality & §14.4(d) \\
      9 & Harmful data bias & §10.2(f), §10.2(g), §10.3 (2) & 30 & Unavailability of stopping functionality & §14.4(e) \\
       \cmidrule{4-6}
      10 & Incomplete data & §10.3 (1) &\multicolumn{3}{l}{\textbf{ Technical and Cybersecurity}} \\
       \cmidrule{4-6}
       
      11 & Irrelevant data & §10.3 (1) & 31 & Insufficient accuracy & §15.1 \\
      12 & Erroneous data & §10.3 (1) & 32 & Lack of robustness & §15.1 \\
      13 & Inappropriate statistical properties of data & §10.3 (2) & 33 & Vulnerability against cybersecurity-related attacks & §15.1, §15.1 (1) \\
      14 & Sensitive information leakage & §10.5(b)& 34 & Lack of resilience against adverse conditions & §15.4 (1) \\
      15 & Sensitive information misuse & §10.5(c)& 35 & Self-Reinforcing Bias Through Feedback Loops & §15.4 (3) \\
      16 & Unauthorized access to sensitive information& §10.5(c)&36 & Confidentiality attacks & §15.5 (3) \\
      17 & Insufficient traceability of AI system functioning & §12.2  & 37 & Exploitation of model flaws & §15.5 (3) \\
       \cmidrule{1-3}
       \multicolumn{3}{l|}{\textbf{Human-AI Interaction}}
       
       & 38 & Model poisoning & §15.5 (3) \\
       \cmidrule{1-3}
      18 & Lack of transparency during operation & §13.1 & 39 & Data poisoning & §15.5 (3) \\
      19 & AI output lacks interpretability & §13.1, §13.3(b) (vii), §14.4(c) & 40 & Adversarial input & §15.5 (3) \\
      20 & Unwanted misuse & §13.1 & & & \\
      \bottomrule
    \end{tabularx}
\end{table}